\documentclass[aoas,preprint]{imsart}

\RequirePackage{amsthm,amsmath,amsfonts,amssymb}
\RequirePackage[authoryear]{natbib}
\RequirePackage[colorlinks,citecolor=blue,urlcolor=blue]{hyperref}
\RequirePackage{graphicx}% uncomment this for including figures
\usepackage{listings}
\usepackage{tabulary}

\startlocaldefs
\def\bp{\mathbf{p}}

\def\br{\mathbf{r}}
\def\bs{\mathbf{s}}

\def\bu{\mathbf{u}}
\def\bv{\mathbf{v}}

\def\bz{\mathbf{z}}

\def\bI{\mathbf{I}}

\def\bP{\mathbf{P}}

\def\bX{\mathbf{X}}

\newcommand{\bfbeta}{\mbox{\boldmath $\beta$}}

\def\diag{\textrm{diag}}

\def\exp{\textrm{exp}}

\endlocaldefs

\begin{document}

\begin{frontmatter}
\title{Bayesian Inverse Reinforcement Learning for Collective Animal Movement}
%\title{A sample article title with some additional note\thanksref{T1}}
\runtitle{Collective Movement Inverse Reinforcement Learning}
%\thankstext{T1}{A sample of additional note to the title.}

\begin{aug}
%%%%%%%%%%%%%%%%%%%%%%%%%%%%%%%%%%%%%%%%%%%%%%
%%Only one address is permitted per author. %%
%%Only division, organization and e-mail is %%
%%included in the address.                  %%
%%Additional information can be included in %%
%%the Acknowledgments section if necessary. %%
%%%%%%%%%%%%%%%%%%%%%%%%%%%%%%%%%%%%%%%%%%%%%%
\author[A]{\fnms{Toryn L. J.} \snm{Schafer}\ead[label=e1]{tls255@cornell.edu}},
\author[A]{\fnms{Christopher K.} \snm{Wikle}\ead[label=e2]{wiklec@missouri.edu}}
\and
\author[B,C]{\fnms{Mevin B.} \snm{Hooten}\ead[label=e3]{hooten@rams.colostate.edu}}
%%%%%%%%%%%%%%%%%%%%%%%%%%%%%%%%%%%%%%%%%%%%%%
%% Addresses                                %%
%%%%%%%%%%%%%%%%%%%%%%%%%%%%%%%%%%%%%%%%%%%%%%
\address[A]{University of Missouri}
\address[B]{U. S. Geological Survey, Colorado Cooperative Fish and Wildlife Research Unit}
\address[C]{Colorado State University}
\end{aug}

\begin{abstract}
%Limited to 200 words.
Agent-based methods allow for defining simple rules that generate complex group behaviors. The governing rules of such models are typically set \textit{a priori} and parameters are tuned from observed behavior trajectories. Instead of making simplifying assumptions across all anticipated scenarios, inverse reinforcement learning provides inference on the short-term (local) rules governing long term behavior policies by using properties of a Markov decision process. We use the computationally efficient linearly-solvable Markov decision process to learn the local rules governing collective movement for a simulation of the self propelled-particle (SPP) model and a data application for a captive guppy population. The estimation of the behavioral decision costs is done in a Bayesian framework with basis function smoothing. We recover the true costs in the SPP simulation and find the guppies value collective movement more than targeted movement toward shelter.
\end{abstract}

\begin{keyword}
\kwd{agent-based model}
\kwd{inverse optimal control}
\kwd{Markov decision process}
\kwd{variational approximation}
\end{keyword}

\end{frontmatter}
%%%%%%%%%%%%%%%%%%%%%%%%%%%%%%%%%%%%%%%%%%%%%%
%% Please use \tableofcontents for articles %%
%% with 50 pages and more                   %%
%%%%%%%%%%%%%%%%%%%%%%%%%%%%%%%%%%%%%%%%%%%%%%
%\tableofcontents

%%%%%%%%%%%%%%%%%%%%%%%%%%%%%%%%%%%%%%%%%%%%%%
%%%% Main text entry area:
\section{Introduction}\label{CH_04:intro}

Understanding individual animal decision-making processes in social groups is challenging. Traditionally, agent-based models (ABMs) of individual interactions are used as building blocks for complex group dynamics \citep{vicsek1995novel, couzin2002collective, scharf2016dynamic, mcdermott2017hierarchical, scharf2018process}. ABMs attempt to recreate what is observed in nature by defining a mechanistic model \textit{a priori}. While the simple individual-based rules lead to complex group dynamics, ABMs suffer from preprogrammed behavior after reaching some equilibrium, challenges to incorporate interactions with habitat, and no notion of memory \citep{ried2019modelling}. The goal of inverse modeling is to instead learn the underlying local rules from observations of sequential behavior decisions \citep{lee2017agent,kangasraasio2018inverse, yamaguchi2018identification}. 

Parameters of ABMs in practice need to be tuned or learned by supervised learning \citep{ried2019modelling,wikle2016hierarchical,hooten2020statistical}. A recent alternative to supervised learning is reinforcement learning (RL). RL is goal-oriented learning from continuous interaction between an agent and its environment \citep{sutton1998introduction}. That is, RL methods learn parameters controlling global behavior by trial and error experiments within the defined environment and local rules. The agents learn preferences by paying costs to (or receiving rewards from) the environment and choose optimal behavior by minimizing the cumulative expected future costs (also referred to as ``costs-to-go''). Similar to difficulty in tuning ABMs, defining the cost function to produce desired long term behavior is challenging \citep{ng2000algorithms, finn2016guided, arora2018survey}. 
%Do I need more refs for this last sentence?

In systems where observations of behavior trajectories can be collected, inverse reinforcement learning (IRL) methods aim to learn the state costs or costs-to-go that governed the observed agents' decisions. \citet{ng2000algorithms} introduced the first IRL algorithms including dynamic programming, which solves a system of equations based on the state transition probabilities and a grid search method for exploring potential state costs that may have generated observed trajectory samples. As surveyed by \citet{arora2018survey}, many more methods have since been developed or adapted to address problems of meaningful size and non-identifiability of the costs. In fact, \citet{ng2000algorithms} showed in general there does not exist a unique solution to the state costs for systems with finite state space which requires subsequent modeling assumptions. The methods can be broadly categorized as maximum margin optimization \citep{ratliff2006maximum}, entropy optimization \citep{ziebart2008maximum}, Bayesian IRL \citep{ramachandran2007bayesian,choi2011map, jin2015inverse, sosic2017bayesian}, and deep learning IRL \citep{wulfmeier2015deep}, with the majority of the methods being applied to Markov decision processes (MDPs). The benefit of Bayesian frameworks to address the non-identifiability of costs in IRL is that they provide a distribution of costs that can generate the observed expert behavior and incorporate prior information to constrain state costs \citep{ramachandran2007bayesian}.

Many of the aforementioned methods parameterize the likelihood by the immediate state costs, because the state cost function is a concise description of the task \citep{ng2000algorithms,ramachandran2007bayesian}. A computational challenge associated with parametrizing the likelihood by the costs is the necessity to solve the forward MDP each iteration as often the likelihood still involves calculating the state costs-to-go. An alternative class of MDP, the linearly-solvable MDP (LMDP) introduced by \citet{todorov2009efficient}, is linear in its solution for the optimal policy and thus, less computationally costly for forward modeling. The LMDP is defined by a set of passive dynamics that describe an agent's state transitions in the absence of state costs or environmental feedback and then the optimal state transitions minimize costs-to-go. Moreover, IRL for LMDPs does not require the forward solution for each iteration as there is a linear relationship between the costs-to-go and immediate state costs. Therefore, inference about immediate state costs can be obtained by transformation of the estimated costs-to-go. As a special case, \citet{dvijotham2010inverse} showed that maximum entropy IRL is the solution to an LMDP with uniform passive dynamics. However, the maximum entropy IRL algorithms are parameterized by the state costs and require the computationally intensive step of solving the forward RL problem \citep{ziebart2008maximum}. \citet{kohjima2017generalized} proposed a Bayesian IRL method for learning state costs-to-go for LMDPs using variational approximation. 

%A popular class of IRL methods is  with a potential function based on the state-action values and feature matching. Bayesian IRL methods also assume a trajectory likelihood with a state-action value potential function resulting in a multinomial likelihood. Various Bayesian IRL implementations include using \textit{maximum a posteriori} estimation with gradient optimization \citep{choi2011map,choi2014hierarchical}, Gaussian Process IRL \citep{jin2015inverse}, and full MCMC sampling \citep{sosic2017bayesian}. Bayesian nonparametric models have been used for reward feature construction \citep{choi2013bayesian} and learning the representation of state information used by the agent \citep{sosic2017bayesian}. 

%All of the mentioned methods are solving IRL in Markov decision processes (MDPs) and methods 

As argued by \citet{ried2019modelling}, an MDP (or LMDP) for collective animal movement is a better model for the system than traditional self-propelled particle (SPP) models \citep{vicsek1995novel}. The MDP incorporates the internal processes of an animal by modeling the behavior as perception (state space), planning (state values), and action (see \citet{hooten2019running} for related individual-level models). Furthermore, the behavior is governed by feedback from the environment (which includes other agents) rather than assuming automatic interaction rules. Few applied examples of IRL for collective animal movement exist in the literature. Exceptions include the application of maximum entropy IRL to flocking pigeons of \citet{pinsler2018inverse} and Bayesian policy estimation of the SPP and Ising models \citep{sosic2017inverse}. 
%Should I generalize to all animal movement?

%Put some novelty here

We present the first application of IRL for collective animal movement using Bayesian learning of state costs-to-go for an LMDP.  As an extension of \citet{kohjima2017generalized}, we reduce the dimension of the state space with basis function approximation, compare variational approximation to MCMC sampling, and consider the multi-agent LMDP. We first demonstrate the modeling framework for a simulation of the \citet{vicsek1995novel} SPP model to illustrate the mechanisms of the LMDP framework in Section \ref{CH_04:vicsek}. In Section \ref{CH_04:guppies}, we use the new methodology to estimate state costs-to-go for collective movement of guppies  (\textit{Poecilia reticulata}) in a tank to infer trade offs between targeted motion and group cohesion. Finally, we discuss the findings and direction for future work in Section \ref{CH_04:disc}. 

%Other examples of applied IRL for animal behavior include \citet{yamaguchi2018identification}. we choose to take the approach of \citet{yamaguchi2018identification} with the LMDP due to computational efficiency.

%Examples of IRL for animal movement collective \citep{sosic2017inverse, pinsler2018inverse}. Examples of IRL for animal movement/trajectory prediction \citep{hirakawa2018can, takemura2019trajectories}. Our work is different in the UQ and use of the LMDP. (although max ent is a special case of LMDP).
%
%What we are doing
%
%roadmap
%LMDP IRL based on the values, but with linearity, can solve for the costs.

%Couzin model is widely used for modeling collective behavior in a forward fashion; zonal type behavior.

% Animal movement and habitat use under collective movement. We want inference, but ABMs or statistical/resource selection models more two-stage/difficult. Resource selection inference examples of memory exist, but maybe not future planning? whereas MDP will consider future actions if $\gamma >0$. Behavior in variable habitats also difficult. How can you tune an ABM for realistic long term behavior?
%
%Furthermore, we define the MDP for a guppy experiment and use the method to get inference.
%
%We demonstrate IRL for guppies. Collective movement in MDP/RL under variable habitat possible as shown in our STAT/proceedings. We aim to show here that inference from IRL is possible under the proposed MDP. Bayesian IRL is our method of choice. Using an entropy framework doesn't provide the desired uncertainty

\section{IRL Methodology}
%let's describe everything in terms of the rewards/values vs. costs/costs-to-go

\subsection{LMDP}\label{CH_04:lmdp}

We focus on the discrete state space LMDP defined by the tuple $(S,\bar{\bP},\gamma,R)$ where $S = \{1,...,J\}$ is a finite set of states, $\gamma \in [0,1]$ is a discount factor, $R: S \rightarrow \mathbb{R}$ is a state cost function, and $\bar{\bP}$ is a $J \times J$ transition probability matrix with elements $\bar{p}_{ij}$ for $i = 1,...,J$ and $j = 1,...,J$ corresponding to the transition from state $i$ to state $j$ under no control (e.g., passive dynamics). We denote an observation from the set of states as $\bs \in  \{1,...,J\}$ and the state cost at state $i$ as $r_i$ for  $i = 1,...,J$ (see Appendix \ref{appendix:notation} for a notational reference).

The policy (e.g., how to choose the next state) of an LMDP is defined by continuous controls, $\bu = \{u_{j} \in  \mathbb{R}; \forall j = 1,...,J\} $, such that the controlled dynamics are expressed as:
    \begin{equation}\label{eq:controlprob}
    	p(s_t = j \vert s_{t-1} = i) = p_{ij}(\bu) \equiv \bar{p}_{ij}\exp(u_{j}),  
    \end{equation}
and the controls are defined to be 0 when the passive transition probability is 0 (i.e., if $ \bar{p}_{ij} = 0$, then $ p_{ij}(\bu) = 0$). The controls, $u_{j}$, are interpretable as the cost the agent is willing to pay to go against the passive dynamics \citep{todorov2009efficient}. For a given policy, the joint costs of the state and control, $l(i,\bu)$, are:
\begin{equation}\label{eq:costs}
	 l(i,\bu) = r_i + KL(\bp_{i}(\bu) \vert \vert \bar{\bp}_{i}),
\end{equation}
where $r_i$ is the immediate state cost for states $i = 1,...,J$ and $KL(\cdot)$ is the Kullback-Leibler (KL) divergence between the controlled transition probability, $\bp_{i}(\bu) = (\bar{p}_{i1}\exp(u_{1}),...,\bar{p}_{iJ}\exp(u_{J}))'$, and passive transition probabilities, $\bar{\bp}_{i} = (\bar{p}_{i1},...,\bar{p}_{iJ})'$. The KL divergence penalty requires the agent to ``pay'' a larger price for behavior that deviates from the passive dynamics \citep{todorov2007linearly}. 

The state costs-to-go, $v_i$, for $i = 1,...,J$, are the discounted sum of future expected costs incurred from beginning in state $i$:
\begin{equation}\label{eq:CTG}
	v_i = l(i,\bu) + E[\gamma \sum_{t=1}^T l(j,\bu)],
\end{equation}	
where the expectation is with respect to the controlled transitions \eqref{eq:controlprob}. The value of $T$ determines whether the problem has finite- or infinite-horizon (e.g., $T < \infty$ or $T = \infty$). A finite-horizon LMDP can be modeled as an infinite-horizon LMDP by assuming the agent remains in the final observed state and incurs no future costs \citep{todorov2007linearly}. Costs-to-go can also be interpreted as relative time to goal completion where a smaller cost-to-go indicates that the agent can reach a desirable state more quickly by transitioning to that state than transitioning to a state with a higher cost-to-go. Based on the definition, there is a recursive relationship between the cost-to-go functions such that \citep{sutton1998introduction,todorov2009efficient}:
\begin{equation}\label{eq:recursion}
	v_i = l(i,\bu) + E[\gamma v_j].
\end{equation}	
The forward problem of the LMDP solves for the optimal set of controls that minimize the cost-to-go and can be expressed by the Bellman optimality equation \citep[e.g.,][]{bellman1957dynamic} for the state costs-to-go, $v_i$, for $i = 1,...,J$: 
\begin{equation}\label{eq:lmdpbell}
v_i = \min\limits_{\bu} \left( l(i,\bu) + \gamma \sum_{\forall j \in \mathcal{S}}  p_{ij}(\bu)  v_j \right),
\end{equation}
where the summation is over the reachable states $j \in S$ as determined by the policy $p_{ij}(\bu)$ for all $j \in S$ (i.e., the expectation in \eqref{eq:CTG} is now expressed as the sum over the discrete distribution defined by \eqref{eq:controlprob}). 
The computational advantage of the LMDP for RL is the Bellman optimality can be solved analytically using the method of Lagrange multipliers for the optimal transition probabilities \citep{todorov2009efficient}:
\begin{equation}\label{eq:pstar}
	p^*(s_t = j \vert s_{t-1} = i) = \frac{\bar{p}_{ij}\exp(-\gamma v_j)}{\sum_{k=1}^J \bar{p}_{ik}\exp(-\gamma v_k)}.
\end{equation}
By substituting equation \eqref{eq:pstar} into the Bellman optimality \eqref{eq:lmdpbell} and exponentiating, the optimal costs-to-go are a solution to an eigenvector problem that is obtained using a power iteration method \citep{todorov2009efficient}, which we demonstrate in Section \ref{CH_04:vicsek}. 

%\subsection{Collective LMDP}
%
%In order to extend the single agent RL to the collective agent setting (or swarm), \citet{sosic2017inverse} does what? Introduces the swarm state and each agent receives a partial observation of the swarm state. 
%
%As in \citet{sosic2017inverse}, we assume that each agent has partial observability of the swarm state (i.e., the collection of locations and headings (angular direction) for all agents in the group). The specifics of each observations are detailed in the data applications below. 

\subsection{Inverse Reinforcement Learning (IRL)}

Assume we observe a collection of sequences of optimal behavioral state trajectories, $\mathcal{D} = \{\mathcal{D}_{1},...,\mathcal{D}_{N}\}$, and $\mathcal{D}_{n} = \{s_{n0},...,s_{nT}\}$, where $s_{nt}$ is the observed state for individual $n$, for $n = 1,...,N$, and time point $t$, for $t = 0,1,...,T$. Then, the observed state transitions are summarized into frequencies, $y_{ij} = \sum_{n=1}^N \sum_{t=1}^T I(s_{nt} = j \vert s_{n(t-1)} = i)$. We assume that each individual operates according to an LMDP with identical parameters, $(S,\bar{\bP},\gamma,R)$, but that the state costs, $R$, and therefore, costs-to-go, $\bv$, are unknown. The likelihood of $\mathcal{D}$ is:
\begin{equation}\label{eq:lmdplik}
	\begin{aligned}
		P(\mathcal{D} \vert \bar{\bP}, \bv) &= \prod_{n=1}^{N} \prod_{t=1}^{T} \prod_{i=1}^{J} \prod_{j=1}^{J} p^*(s_{nt} = j \vert s_{n(t-1)} = i), \\
		&= \prod_{i=1}^{J} \prod_{j=1}^{J} \left(\frac{\bar{p}_{ij}\exp(-\gamma v_j)}{\sum_k \bar{p}_{ik}\exp(-\gamma v_k)}\right)^{y_{ij}},
	\end{aligned}
\end{equation}
for all individuals $n = 1,...,N$, times points $t = 1,...,T$, transitions from state $i \in S$ to state $j \in S$ and the second equality is based on the optimal transitions \eqref{eq:pstar}. We express the costs-to-go vector, $\bv = (v_1,...,v_J)'$, as a linear combination of features in the  $J \times n_b$ matrix $\bX$ with unknown weights $\bfbeta$ (e.g., $\bv = \bX\bfbeta$). We estimate the weights in a Bayesian framework by assuming the following hierarchical prior:
\begin{equation}\label{eq:lmdpprior}
\begin{aligned}
	\bfbeta &\sim \text{N}\left(\boldsymbol{0}, \frac{1}{\tau} \bI_{n_b} \right), \\
	 \tau &\sim \text{Gamma}(0.1,0.1),
\end{aligned}
\end{equation}
where $\boldsymbol{0}$ is an $n_b$-dimensional vector of zeroes and the parameters are estimated using MCMC sampling and variational approximation with the statistical platform Stan using the R package \textit{rstan} \citep{carpenter2017stan,stan2020}. For the MCMC sampling, we used the Hamiltonian Monte Carlo with no-U-turn sampler \citep[e.g.,][]{hoffman2014nuts}, which is the default algorithm in Stan. For variational inference, Stan assumes a Gaussian approximating distribution on a transformation of the parameters to a continuous domain \citep{kucukelbir2015automatic}. We provide brief definitions of the algorithms and the Stan code in an online supplement. Note that the costs-to-go are only estimable up to a constant due to the exponential in \eqref{eq:lmdplik} and therefore all resulting mean costs-to-go functions are shifted to have a minimum value of 0, which typically corresponds to a terminal state or a state in which an agent incurs no cost indefinitely \citep{todorov2009efficient}.

%We use variational Bayesian (VB) inference for inverse reinforcement learning with an LMDP as described in \citet{kohjima2017generalized}. The choice of VB is due to the large parameter space (the model is overparameterized)

%A challenge with collective/multi-agent RL is there is no true model of the environment (aka passive dynamics). Need to estimate the passive dynamics. However, the model is ill-posed. So inference restricted to relative costs/probability.

%\subsubsection{Basis Functions}
%
%The state space is described as the discretization of the target misalignment, local misalignment, and target distance. Therefore, it is natural to assume that the costs-to-go for states that are ``nearby'' will be similar and therefore we can reduce the dimension of the parameter/state space. We chose to reduce the dimension by using a smaller number of basis functions. The basis function methods we explored were radial basis functions with means estimated by k-means on the data and regular Gaussian basis functions \citep{mangion2020frk}.

\section{SPP LMDP}\label{CH_04:vicsek}

We illustrate the LMDP for collective movement using the \citet{vicsek1995novel} SPP model. The SPP model is an ABM for flocking behavior in which collective behavior is induced by an agent assuming the mean direction of agents within its neighborhood \citep{vicsek1995novel}. Therefore, the rule governing the agent's behavior is based on the assumption that a collective agent will travel in the same direction as the group.  

We consider the dynamics of the SPP model for agents $n = 1,...,N$ as formulated by \citet{sosic2017inverse} for the direction $ \theta_{nt}$ and location $(x_{nt}, y_{nt})$ as:
\begin{equation}\label{eq:vicsek}
	\begin{gathered}
		\theta_{n(t+1)} = \langle \theta_{nt} \rangle_{\rho} + \epsilon_{nt}, \quad \epsilon_{nt} \sim \text{N}(0,\sigma^2), \\
		x_{n(t+1)} = x_{nt} + v_{nt} \cdot \text{cos}(\theta_{nt}), \\
		y_{n(t+1)} = y_{nt} + v_{nt} \cdot \text{sin}(\theta_{nt}),
	\end{gathered}
\end{equation}
where the agent heads in the mean direction, $\langle \theta_{nt} \rangle_{\rho}$, of other agents including itself within a neighborhood of radius $\rho$ with a speed of $v_{nt}$. The \textit{local misalignment} of an agent is the difference between the mean neighborhood direction and the agent's direction, $\langle \theta_{nt} \rangle_{\rho} -   \theta_{nt}$. \citet{sosic2017inverse} formulated the SPP model as an MDP with 13 discrete actions corresponding to turning angles,  $\phi \in [-60^\circ, -50^\circ, ..., 60^\circ]$ and a discrete state space defined by a grid of local misalignment values. The local misalignment grid was defined by $J = 36$ equally sized bins of $10$ degrees with centers $\bs = (\pm180^\circ, -170^\circ, ..., 170^\circ)'$. An agent of the SPP MDP chooses a turning angle, $\phi_{nt}$, given the observation of local misalignment bin, $s_{nt} = \sum_{s_i \in \bs} s_i * \textrm{I}(s_i -  5^\circ \leq \langle \theta_{nt} \rangle_{\rho} -   \theta_{nt} \leq s_i + 5^\circ)$ where $\textrm{I}(\cdot)$ is an indicator function. The distribution of the next direction given the turning angle is $\theta_{n(t+1)} \vert \phi_{nt} \sim \text{N}(\theta_{nt} + \phi_{nt}, \sigma^2)$. In our simulation, we assumed a constant velocity of $1$, (i.e., $ v_{nt} = 1 \ \forall \ n =1,...,N; t = 1,...,T$), fixed interaction radius $\rho = 0.1$, and turning angle standard deviation of $10$ degrees (i.e., $\sigma = 10^\circ$).  All angular differences were calculated with the two argument arc-tangent function.

We defined the state cost function, $R$, as:
\begin{equation}\label{eq:SPPcosts}
	r_i =
  \begin{cases}
                                   0 & \text{if $ s_i  = 0^\circ $} \\
                                   2.5 & \text{if $\vert s_i \vert \leq 15^\circ $} \\
                                   4 & \text{if $\vert s_i \vert \leq 25^\circ $} \\
  5 & \text{otherwise}
  \end{cases},
\end{equation}
where the state $s_i \in \bs$, $i =1,...,J$, corresponds to the center of the local misalignment bin. The costs were chosen based on the results of \citet{sosic2017inverse}, which estimated the lowest cost for $s_i  = 0^\circ$ and monotonically increasing costs with increasing  $\vert s_i \vert$ by applying an inverse model to trajectories generated by \eqref{eq:vicsek}. Additionally, the costs were constrained in magnitude such that $\exp(-r_i)$ was not numerically 0 \citep{todorov2009efficient}. 

%In order to generate behavior according to the SPP \eqref{eq:vicsek} under the MDP formulation, the distribution over the next direction needs to be centered on  $\langle \theta_{nt} \rangle_{\rho}$. This corresponds to an MDP policy in which an agent chooses the turning angle that minimizes the current local misalignment. 

\begin{figure}[htp]
\centering
\includegraphics[width=\textwidth,height=\textheight,keepaspectratio]{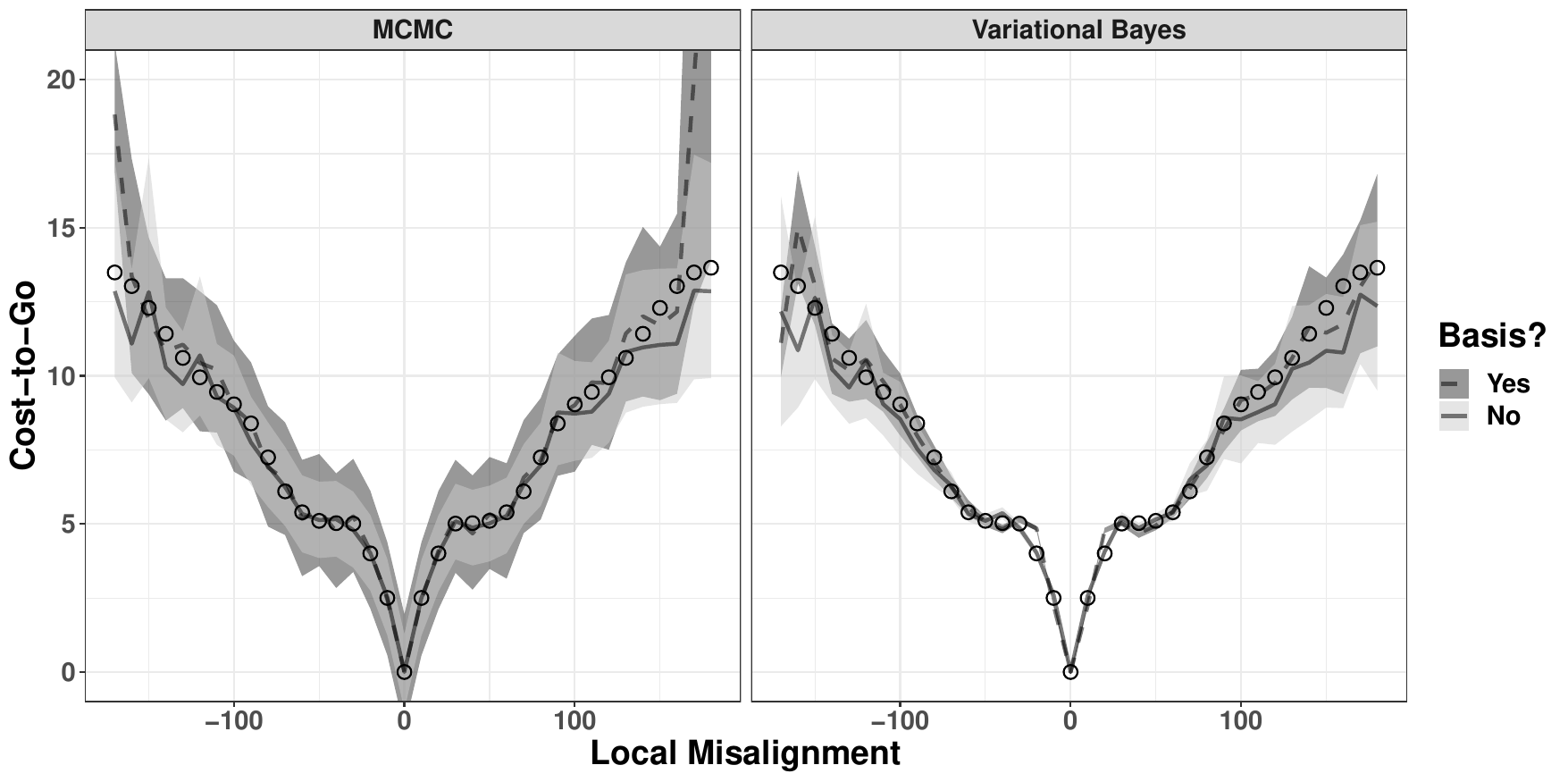}
\caption{Estimated cost-to-go function for a  \citet{vicsek1995novel} SPP model using LMDP IRL for Bayesian MCMC sampling and variational approximation under known passive dynamics. The models either used Gaussian basis functions (dashed lines) or independent state parameters (solid lines). The shaded regions correspond to the 95\% C.I. The open circles is the true cost-to-go function calculated from equation \eqref{eq:ziter}. The mean and true cost-to-go functions were shifted to have minimum 0.}
\label{fig:vicsekcosts}
\end{figure}

To embed the MDP of \citet{sosic2017inverse} into the LMDP framework as outlined by \citet{todorov2007linearly}, we summed over the turning angle action space to derive the transition probabilities for changes in local misalignment states at time $t$ and $t+1$. We assumed agents synchronously chose their next state so the group orientation does not depend on the order of agents' decisions. The turning angle was therefore equivalent to the change in state (i.e., the difference in states was the difference in directions); this implied a continuous transition distribution for the next state given the turning angle, $s_{n(t+1)} \vert \phi_{nt} \sim \text{N}(s_{nt} + \phi_{nt}, \sigma^2)$, which can be discretized to provide a transition probability function over the discrete grid defined by $\bs$. The LMDP passive state transition probabilities were constructed by summing over the conditional transition probabilities given the discrete turning angles, $\phi \in [-60^\circ, -50^\circ, ..., 60^\circ]$, of the MDP.  Therefore the passive dynamics between the discrete grid cell centers $s_{i}$ and $s_{j}$ are a discretization of a mixture of normal distribution functions:
\begin{equation}\label{eq:pbar}
	\begin{gathered}
		\bar{p}(s_{j} \vert s_{i}) \propto  \sum_{\phi \in [-60^\circ, -50^\circ, ..., 60^\circ]} \Phi\left(\frac{s_{j} - s_{i} - \phi + 5^\circ}{10^\circ}\right) - \Phi\left(\frac{s_{j} - s_{i} - \phi - 5^\circ}{10^\circ}\right),
	\end{gathered}
\end{equation}
where $\Phi$ is the standard normal cumulative distribution function and the discretization length $5^\circ$ was determined by the half-length of the state grid cells. The passive dynamics were then normalized to have row sums equal to one (i.e., $\sum_{j \in S} \bar{p}(s_{j} \vert s_{i}) = 1$). Lastly, as stated in Section \ref{CH_04:lmdp}, the true costs-to-go can be calculated as the solution to an eigenproblem. The SPP LMDP setup defines an infinite-horizon problem without an absorbing state (i.e., $\bar{p}_{ii} \neq 1$ for any $i = 1,...,J$) so we choose to consider the average cost LMDP defined by the following system of equations:
\begin{equation}\label{eq:ziter}
	\bz = \frac{1}{\lambda} \diag(\exp(-\br))\bar{\bP}\bz,
\end{equation}
where $\bz = \exp(-\bv)$ is a $J$-dimensional vector referred to as the desirability function, $ \diag(\exp(-\br))$ is a $J \times J$ diagonal matrix with the elementwise exponentiation of the negative state costs, $\exp(-\br) = (\exp(-r_1),...,\exp(-r_J))'$, on the main diagonal, $\bar{\bP}$ is the $J \times J$ passive transition probability matrix, $\lambda$ is the principal eigenvalue of $[ \diag(\exp(-\br))\bar{\bP}]$ and $-\text{log}(\lambda)$ corresponds to the average cost of each time step \citep[see supplementary information in][]{todorov2009efficient}. The scaling by the largest eigenvalue allows for numerical stability in estimation. The system of equations is solved by initializing the vector $\bz$ to all ones, $\bz = \mathbf{1}$, and repeatedly multiplying by $[\frac{1}{\lambda} \diag(\exp(-\br))\bar{\bP}]$ until convergence. This method is referred to as Z-iteration in the LMDP literature \citep{todorov2009efficient}. When applied to the SPP example here, the true cost-to-go function is symmetric about $0^\circ$ with larger relative differences between states near $0^\circ$ than states with local misalignment greater in absolute value than $25^\circ$ (Figure \ref{fig:vicsekcosts}). Because the states with local misalignment values greater in absolute value than $25^\circ$ have the same immediate cost \eqref{eq:SPPcosts}, the differences are related to the average number of time steps it takes an agent to be able to turn toward the group as defined by the passive dynamics; the passive dynamics do not allow an agent to turn more than $90^\circ$ in one step.

We simulated from the calculated optimal policy with 200 agents for 100 time points and calculated the state transition frequencies using the following algorithm:
\begin{enumerate}
	\item Initialize $(x_{n0},y_{n0},\theta_{n0})$ and calculate local misalignment to determine grid cell $s_{n0}$ for $n = 1,...,200$. 
	\item Repeat the following for $t = 1,...,100$ synchronously for $n = 1,...,200$:
	\begin{enumerate}
		\item \label{itm:second} Sample next local misalignment from $p^*( \cdot \vert s_{n(t-1)} = i)$.
		\item \label{itm:third} Calculate turning angle, $\phi_{nt}$, as difference between $\theta_{n(t-1)}$ and \ref{itm:second}.
		\item Update location $(x_{nt},y_{nt})$ according to \eqref{eq:vicsek}, $\theta_{nt} = \theta_{n(t-1)} + \phi_{nt}$, and local misalignment $s_{nt}$.
	\end{enumerate}
\end{enumerate}

We estimated the costs-to-go with full MCMC sampling and variational approximation for comparison. Additionally, we estimated the costs-to-go for each state separately, (e.g., $\bX = \text{I}$), and with Gaussian basis functions with centers on every other grid cell to reduce the state dimension by a factor of 2. 

From Figure \ref{fig:vicsekcosts}, it is evident that all modeling scenarios estimated the relative true costs-to-go and the estimates from MCMC sampling capture more uncertainty than those from variational approximation. It appears the uncertainty of the estimates increases with an increase in local misalignment and the difference is more apparent for the variational approximation. This pattern generally reflects the amount of data; there were more transitions to states with smaller local misalignment. Furthermore, there were no transitions to states in grid cells centered on $-170^\circ, -150^\circ, 170^\circ, \pm180^\circ$ misalignment. 

\begin{figure}[htp]
\centering
\includegraphics[width=\textwidth,height=\textheight,keepaspectratio]{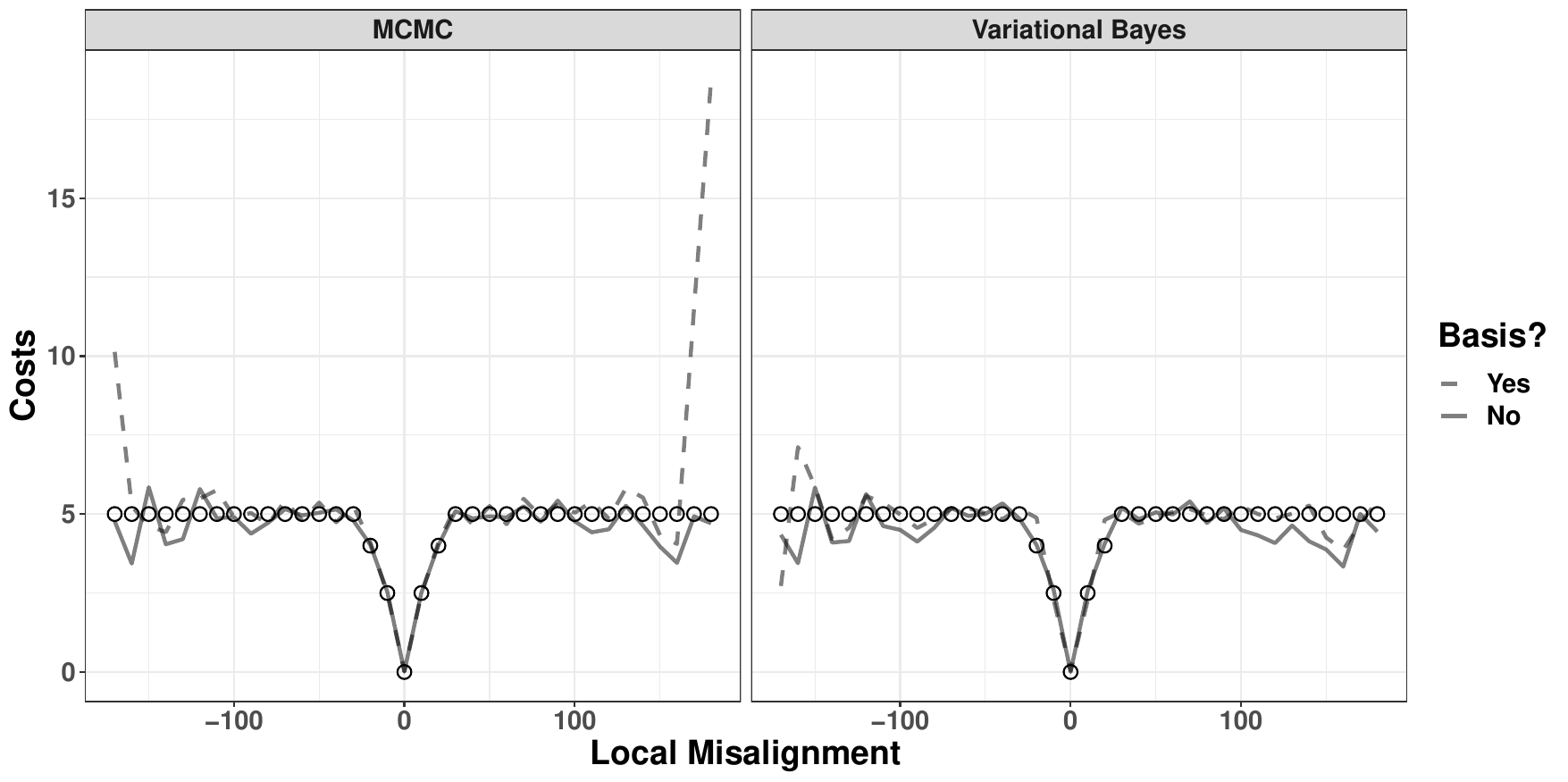}
\caption{Estimated state costs for a  \citet{vicsek1995novel} SPP model using LMDP IRL for Bayesian MCMC sampling and variational approximation under known passive dynamics. The models either used Gaussian basis functions (dashed lines) or independent state parameters (solid lines). The open circles are the true state costs from \eqref{eq:SPPcosts}.}
\label{fig:vicsekstatecosts}
\end{figure}

The LMDP framework for IRL allows for efficient estimation of the state costs $r_i$ from the estimation of the cost-to-go by rearranging \eqref{eq:ziter}:
\begin{equation}\label{eq:estimatecosts}
	r_i = \text{log}(\lambda) + v_i + \text{log}\left(\sum_{j} \bar{p}_{ij} \exp(-v_j)\right),
\end{equation}
for $i = 1,...,J$. Figure \ref{fig:vicsekstatecosts} shows that the estimated costs from the mean cost-to-go functions in Figure \ref{fig:vicsekcosts} generally match the arbitrary state costs defined in \eqref{eq:SPPcosts}. 

For the MCMC estimation with bisquare basis functions, there is an increase in cost-to-go and uncertainty at the boundaries. The obvious spike at $180^\circ$ is the cost-to-go for the state defined by local misalignment less than $-175^\circ$ and greater than $175^\circ$. The Gaussian basis functions were not defined on a circle, but rather the continuous real line and could be contributing to the lack of smoothness near the boundary. Additionally, some flexibility is lost in estimation by reducing the dimensionality of the state space with the basis functions. 

\section{Guppy Application}\label{CH_04:guppies}

\begin{figure}[htp]
\centering
\includegraphics[width=\textwidth,height = \textwidth, keepaspectratio]{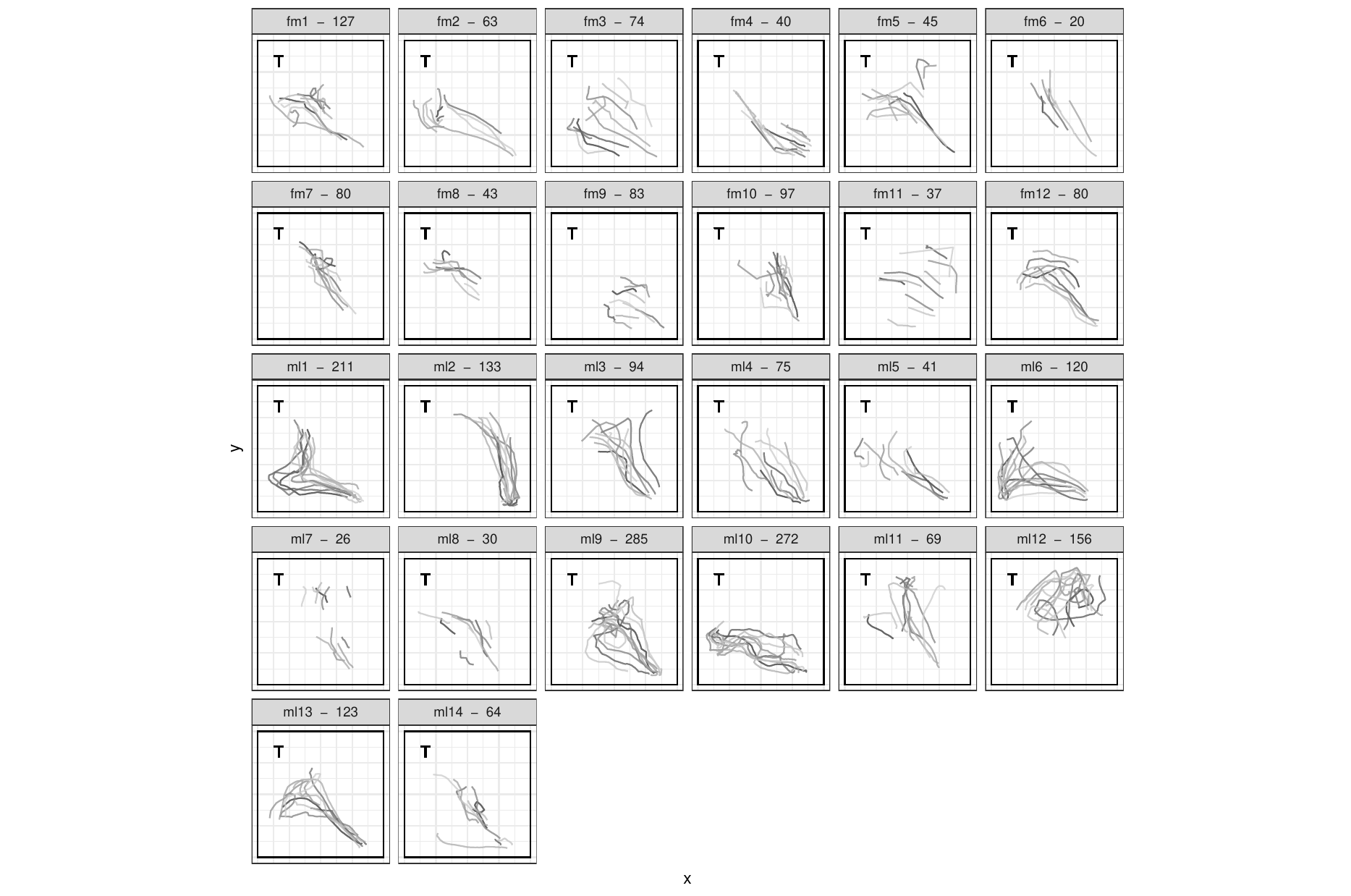}
\caption{Trajectories of all 26 experiments of groups of 10 guppies in a tank. The target is located at the point marked ``T.'' The panel labels are ``experiment number" -  sample size. Experiments in first two rows are female (fm) groups and rows 3-5 are male (ml) groups.}
\label{fig:guppytrajs}
\end{figure}

We used the data publicly available from \citet{bode2012distinguishing} on an experiment involving a captive population of guppies (\textit{Poecilia reticulata}). Groups of 10 same sex guppies were filmed from above in a square tank with one corner containing gravel and shade, which is defined by a point. The shaded corner provided shelter and is hypothesized to be attractive to the guppies. The guppies were released in the tank in the opposite corner. \citet{bode2012distinguishing} determined positions from video tracking software at a rate of 10 frames per second. The data made available by \citet{bode2012distinguishing} consist of movement trajectories truncated to the time points when all individuals were moving until one guppy reached the shaded target area; the range of time series was 20 - 285 frames. There were 26 experiments with 12 experiments consisting of all females and 14 males (Figure \ref{fig:guppytrajs}). We used trajectories from all experiments to estimate the cost-to-go functions based on movement headings.

%The public data set only defines the target by a point. 

We defined an LMDP for the guppy trajectories with a discrete state space of local misalignment and target misalignment. Local misalignment was defined as in Section \ref{CH_04:vicsek} and target misalignment was defined as the difference between the current heading and the direction to the target point. We rescaled all pixel locations to the unit square and calculated the local misalignment between an individual and all other individuals. The assumption of interaction with all other agents is reasonable as the movement was bounded and there were no visual obstructions outside the target area \citep{bode2012distinguishing}. The two misalignment states were discretized using the same $36$ bins of length $10^\circ$ as in the previous section resulting in a discretized grid of $J=36 \times 36$ states. We assume a fixed discount factor of 1 (i.e., all future costs/rewards are not discounted). For state transitions, we assumed synchronous updates as in the SPP simulation. Across the 26 experiments, there were 7,816 unique state transitions. Estimation of parameters in the guppy application was done by variational approximation due to the size of the state space. 

\begin{figure}[htp]
\centering
\includegraphics[width=\textwidth,height = \textwidth, keepaspectratio]{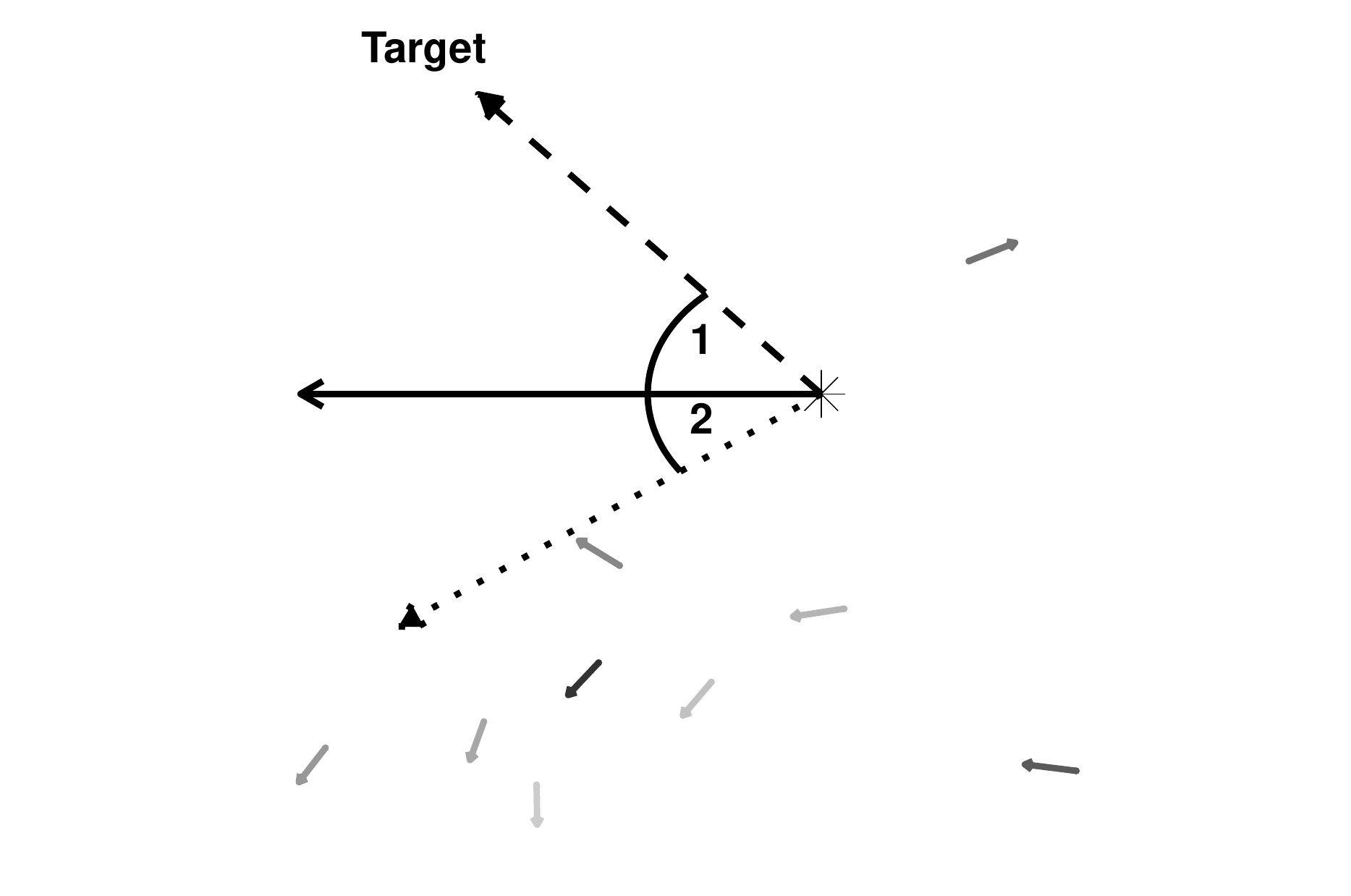}
\caption{An illustration of the state for a guppy at the location marked by a star. The large solid arrow indicates the guppy's heading. The dashed arrow indicates the straight-line direction from the guppy's location and the angle marked by 1 is the guppy's \textit{target misalignment} (i.e., how much the guppy should change its current heading to face the target). The dotted line indicates the current average group direction as calculated from all of the solid arrows (including itself) and the angle marked by 2 is the guppy's \textit{local misalignment} (i.e., how much the guppy should change its current heading to head in the average direction). The angles labelled 1 and 2 would have opposite signs where turning angle 1 would result in a turn to the right and turning angle 2 would result in a turn to the left.}
\label{fig:stateexample}
\end{figure}

\begin{figure}[htp]
\centering
\includegraphics[width=\textwidth,height=\textheight,keepaspectratio]{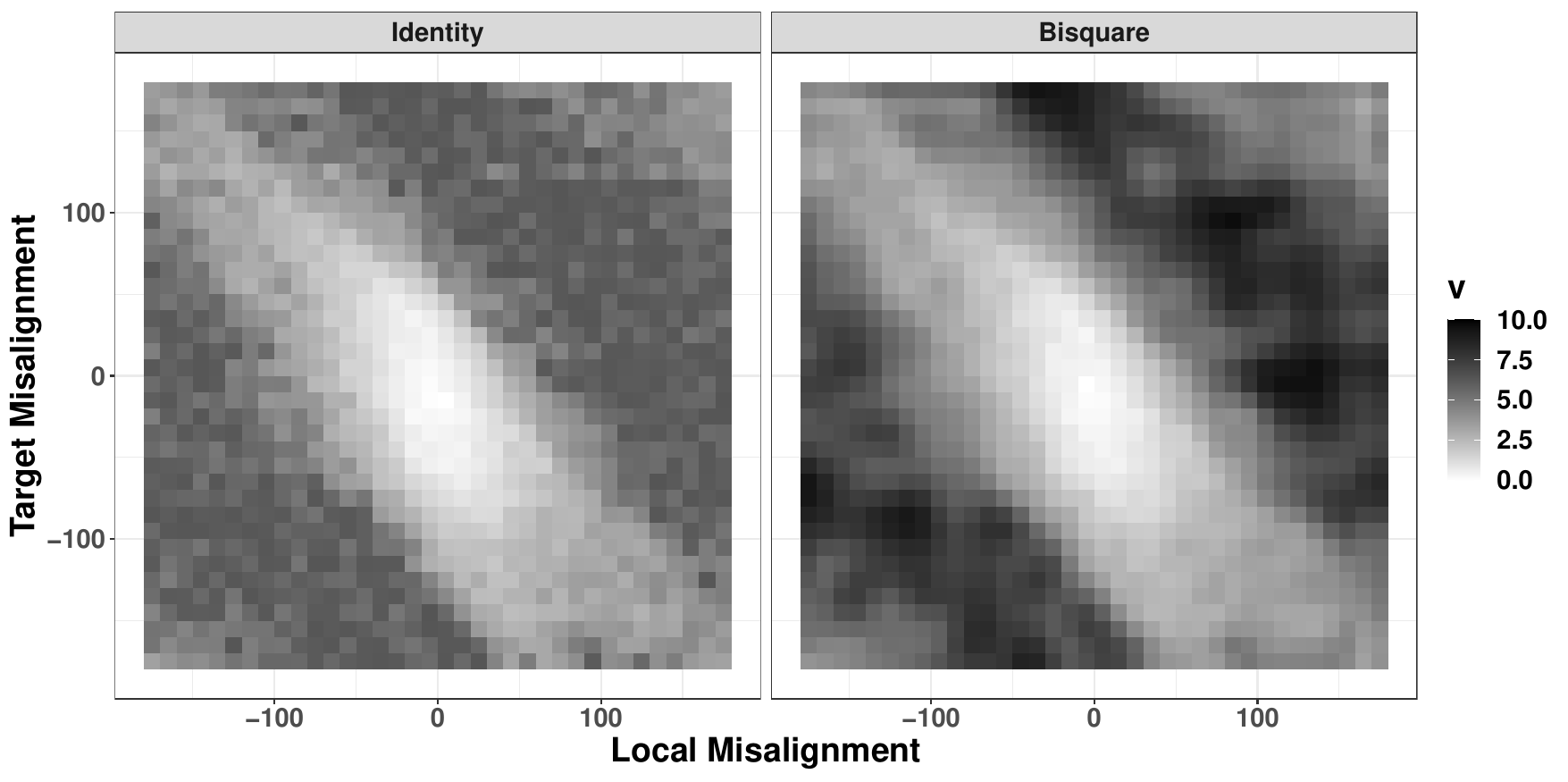}
\caption{Variational posterior mean costs-to-go for the guppy experiments for a gridded state space of target and local misalignment across two sets of features: full (identity matrix) and bisquare basis functions. The passive dynamics are assumed to be discrete uniform and the mean estimated costs have been shifted to have a minimum of 0. A lower value associated with the legend for \textbf{v} indicates states with lower costs-to-go and therefore states to which the guppies are estimated to choose to transition.}
\label{fig:guppybases}
\end{figure}

For the first set of estimated costs-to-go, we assumed the passive dynamics to be discrete uniform, (e.g., $\bar{p}_{ij} \propto 1$ for all $i,j = 1,...,J$). The features, $\bX$, considered were the identity matrix and 819 multiresolution bisquare basis functions generated uniformly within the gridded state space by the R package \textit{FRK} \citep{mangion2020frk} referred to as ``Identity'' and ``Bisquare'' respectively in Figure \ref{fig:guppybases}. The results shown in Figure \ref{fig:guppybases} show a similar pattern among feature matrices with the bisquare basis functions providing more smoothing across the state space. In general, the results suggest the guppies perceived less cost for aligning with other guppies as the low costs-to-go in the center of the figure are concentrated around $0^\circ$ and there is more flexibility in target alignment as the low costs-to-go have more spread along the target misalignment axis. When comparing the two feature matrices (Figure \ref{fig:guppybases}), there is more contrast between the estimated cost-to-go function values estimated with bisquare basis functions than the identity matrix that is likely attributable to the dimension reduction. 

%The guppy experiment can be considered a first exit MDP, because the experiment has a defining end criterion. The first exit problem can be modeled as infinite-horizon by assuming the last state is an absorbing state with no costs (=infinite reward?). The first exit condition is a on the swarm state (at least one guppy is in the shaded area). 

%Unlike \citet{kohjima2017generalized}, we assume we know the passive dynamics of the system for simplicity and it is a uniform random walk bounded within the unit square.

\begin{figure}[htp]
\centering
\includegraphics[width=\textwidth,height=\textheight,keepaspectratio]{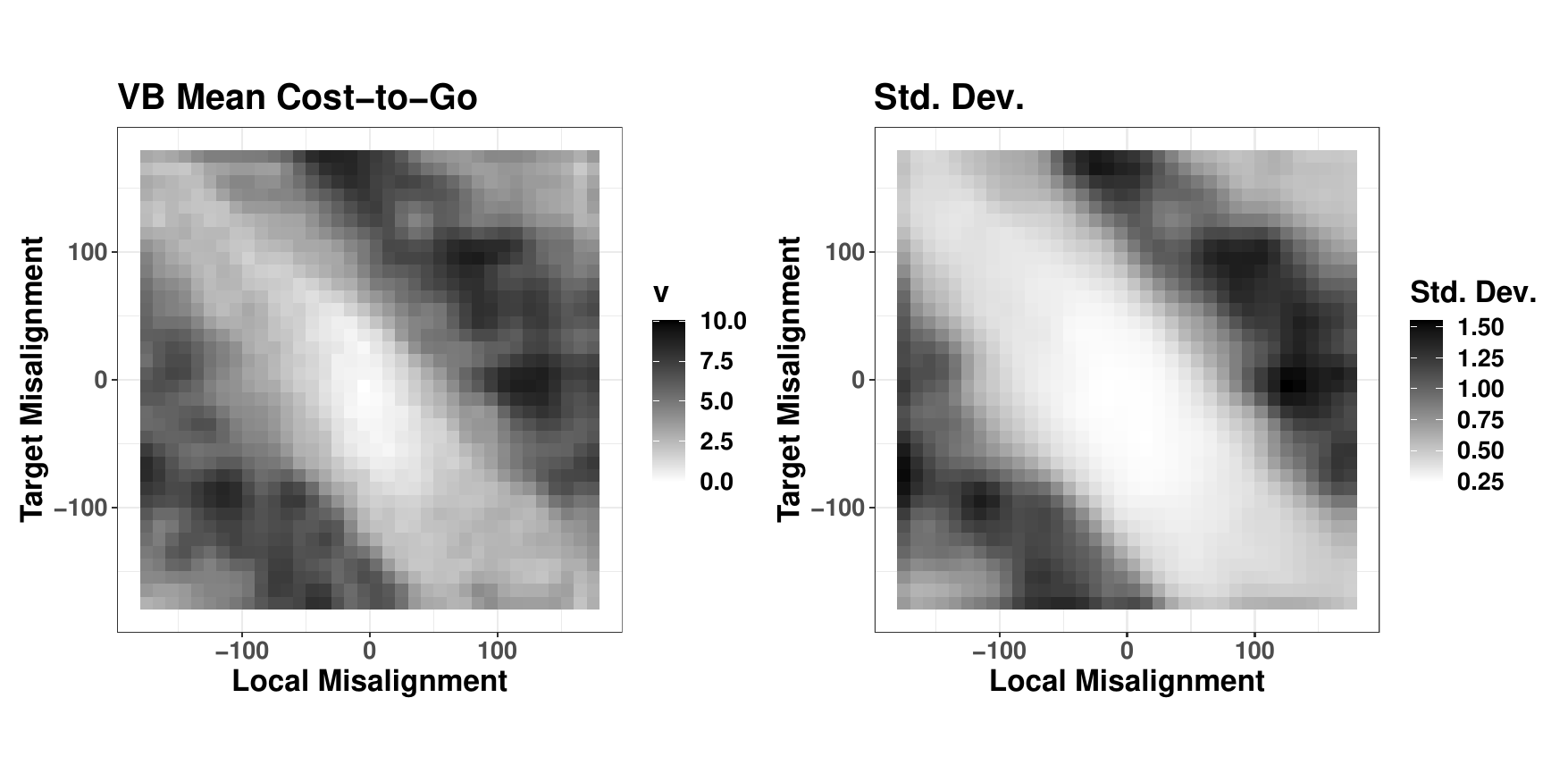}
\caption{Variational posterior mean costs-to-go (left panel) and standard deviations (right panel) for the guppy experiments for a gridded state space of target and local misalignment with passive dynamics assumed to be a normal random walk and bisquare basis functions. The mean estimated costs-to-go have been shifted to have a minimum of 0. A lower value associated with the legend for \textbf{v} indicates states with lower costs-to-go and therefore states to which the guppies are estimated to choose to transition.}
\label{fig:guppyRW}
\end{figure}

To assess the sensitivity to the assumed passive dynamics, we estimated the costs-to-go under a set of passive dynamics corresponding to an independent, normal random walk on the gridded state space with standard deviation $90^\circ$. The standard deviation was chosen to be large enough to ensure all non-zero transition probabilities to use all of the observed data. The variational posterior mean and standard deviation for the costs-to-go are shown in Figure \ref{fig:guppyRW}. Comparing to the previously estimated states, the variational posterior mean cost-to-go functions are similar. The variational posterior standard deviations reflect the pattern of observed frequencies with states more frequently observed having smaller uncertainty.   

In Figures \ref{fig:guppybases} and \ref{fig:guppyRW}, the diagonal pattern can be attributed to the corners appearing far when plotted in the 2-D plane, but are close together in circular space so they have similar costs-to-go. 

\begin{figure}[htp]
\centering
\includegraphics[width=\textwidth,height=\textheight,keepaspectratio]{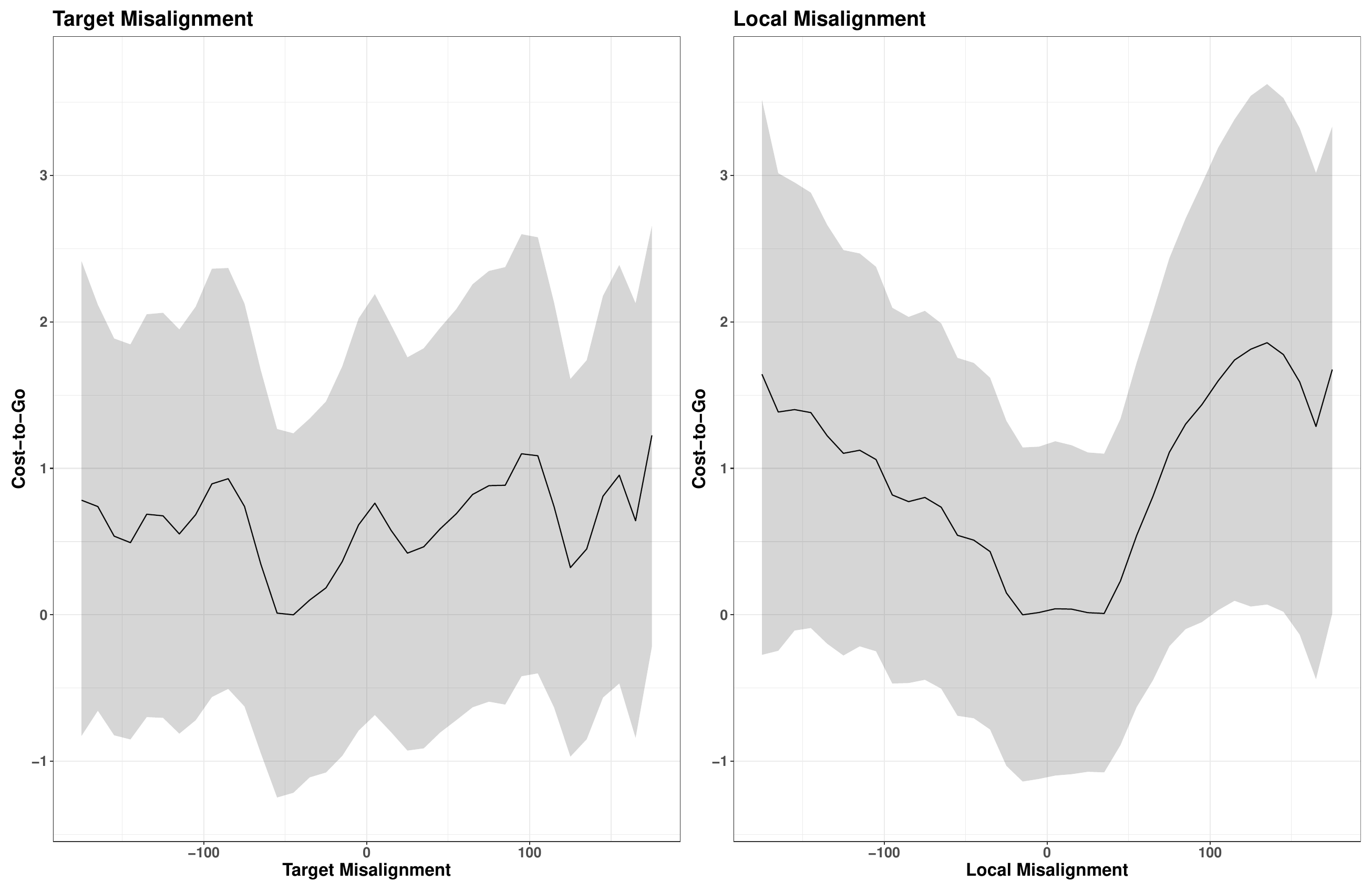}
\caption{Marginal costs-to-go of target and local misalignment for the guppy experiments for a gridded state space of target and local misalignment with passive dynamics assumed to be a normal random walk and bisquare basis functions. The mean estimated costs have been shifted to have a minimum of 0. The lower costs-to-go increase the probability of transitioning into that state.}
\label{fig:guppymarginals}
\end{figure}

The marginal costs-to-go based on the estimates in Figure \ref{fig:guppyRW} are shown in Figure \ref{fig:guppymarginals}, where the costs-to-go are calculated as the mean across all values of the other state variable and shifted to have a minimum of 0. There is evidence of collective alignment as shown in the local misalignment costs-to-go function due to the minimum cost occurring at $0^\circ$ with gradual increase as misalignment increases in absolute value. Furthermore, the guppies appear to perceive local misalignments from $-15^\circ$ to $45^\circ$ as equally optimal, which can be contrasted with the sharp dip in cost-to-go for $0^\circ$ local misalignment in the SPP simulation Figure \ref{fig:vicsekcosts}. The dip in the target misalignment cost-to-go function corresponds to the grid cells defined by $-55^\circ$ to $-45^\circ$ and $-45^\circ$ to $-35^\circ$, suggesting it is less costly to approach the upper corner with the target  $55^\circ$ to $35^\circ$ to the right. From inspection of the observed data shown in Figure \ref{fig:guppytrajs}, it appears many of the guppies moved across the tank to the left first, which would require a right turn to decrease the target misalignment. A symmetry constraint could be applied to the costs-to-go by considering the absolute target alignment if it were assumed to be equally costly to approach from the right or left. Simulations of an example of guppy movement and estimated decision making from the costs-to-go in Figure \ref{fig:guppyRW} can be found at \href{https://github.com/schafert/inverse-irl-guppy}{https://github.com/schafert/inverse-irl-guppy}.

\section{Discussion}\label{CH_04:disc}

Collective motion from generative local interaction rules limit possible behavior, but the (L)MDP framework extends the definition of the agent to include perception and internal processes \citep{ried2019modelling}. By estimating the state costs-to-go or value functions, system specific local rules can be estimated. 

Our analysis of the captive guppy populations confirms previous works that find evidence of social interactions between individuals \citep{bode2012distinguishing,russell2016dynamic,mcdermott2017hierarchical}. However, instead of defining a set of behavioral rules \textit{a priori}, we estimated the decision-making mechanisms. Our results suggested the captive guppies value collective movement more than targeted movement toward shelter which was not previously explored by \citet{russell2016dynamic} or \citet{mcdermott2017hierarchical}. Furthermore, the behavioral mechanisms determined by the cost-to-go functions were non-linear and non-symmetric.

In general, our inference is constrained to relative differences in costs-to-go. This is similar to the estimation of relative selection probabilities in animal resource selection modeling \citep{hooten2017animal,hooten2020animal} and therefore IRL can still provide useful inference. However, the SPP simulation demonstrated the ability to recover the magnitude of the true state costs and the true magnitude of the cost-to-go function. 

%Additionally, a different model parameterization which constrains the 

% Defining states still hard and may be ill-posed?- may be use HMM to estimate latent states and either use or guide state definition (two-stage or joint model) \citep{kang2018data}. It depends on goal - if want to simulate, then we don't need to know what the states are, but inference is always better with defined states (not latent).

It may be possible to improve the inference for the guppy data by relaxing the assumptions, estimating passive dynamics, and expanding the state space to include other features. We tested sensitivity of inference to choice of passive dynamics with two simple models. We did not detect a substantial difference, but for full quantification of uncertainty, joint estimation of passive dynamics could be considered. In future work, estimation of the passive dynamics parameters such as the random walk variance may be helpful. Additionally, the state space could include features based on physical distance to assess hypotheses about zonal collective movement which is a primary feature of collective movement ABMs \citep[e.g.,][]{couzin2002collective}. A feature based on distance to target could similarly allow for interaction with the target to vary with location to relax the assumption that an individual interacts with the target in the same manner everywhere. 

In the SPP simulation and guppy application, we assumed a discount factor of 1 which may be realistic for trajectories from such a short time frame. For observations spanning longer periods of time, it would be more realistic to assume there is some loss of memory about past states which would correspond to a discount factor less than 1. Additionally, the discount factor can also be interpreted as the degree to which agents behave optimally \citep{choi2014hierarchical}. It might be expected that observations from animals in the wild are subject to more stochasticity than experimental settings and therefore do not always behave optimally. 

Another modeling choice was the grid size of the discrete state space. There is a precedence for discrete state spaces in animal movement modeling \citep{hooten2010agent, hanks2015continuous}. Biologically, the assumption of a discrete state space assumes the individual perceives values within the range of the bin as equally costly. For the guppy experiments, the evaluation of the state at a $10^\circ$ resolution could be adjusted given expert opinion or model based estimates of navigational ability \citep{flemming2006how}. 

There exist several avenues for extensions of the model to accommodate the behavior of free-ranging animals. First, alternative animal movement models could be proposed for the passive dynamics \citep{hooten2017animal}. Second, covariates could be included on either the costs-to-go or passive dynamic models which would allow for heterogeneous decision making. Another interesting avenue would be to explore the methodology related to subtask completion for LMDPs \citep{earle2018hierarchical}. For example, a group of animals on the landscape navigating to a destination may require the completion of subtasks such as traversing wildlife corridors.

%Future include observation error in other data sets?
%
%The state values are only estimable up to a constant and inference is limited to relative values. Exploring other model specification to give explicit inference...

%I actually think the Vicsek model with estimated passive dynamics is overfit to the expert data observed. This shows that mis-specifying the passive dynamics can have exact inferential consequences, but overall pattern is similar.

%In general, states have lower costs when there are multiple ways to reach the goal from that state \citep{todorov2009efficient}
%
%There are many avenues for future applications of LMDPs for collective animal movement. Due to the linearityTreating the things as subtasks in animal movement, e.g., must pass through a corridor.
%
%Extend to individual behavior/personality, i.e., relax assumptions about equal state costs.
%
%note that features on the state costs (or rewards) can be done, but lose the computational efficiency. \citet{dvijotham2010inverse} suggest it may be more efficient to estimate the effects of features on rewards after estimating the costs-to-go using the relationship in \eqref{eq:costs}.

%%%%%%%%%%%%%%%%%%%%%%%%%%%%%%%%%%%%%%%%%%%%%%
%% Single Appendix:                         %%
%%%%%%%%%%%%%%%%%%%%%%%%%%%%%%%%%%%%%%%%%%%%%%
%\begin{appendix}
%\section*{???}%% if no title is needed, leave empty \section*{}.
%\end{appendix}
%%%%%%%%%%%%%%%%%%%%%%%%%%%%%%%%%%%%%%%%%%%%%%
%% Multiple Appendixes:                     %%
%%%%%%%%%%%%%%%%%%%%%%%%%%%%%%%%%%%%%%%%%%%%%%
\begin{appendix}

\section{LMDP Notation}\label{appendix:notation}

The following is a table of LMDP notation used throughout the manuscript in order of appearance:

\settowidth\tymin{$p^*(s_t = j \vert s_i = i)$}
\setlength\extrarowheight{2pt}
\begin{tabulary}{\textwidth}{L|L}
	\textbf{Symbol} & \textbf{Definition}\\
	\hline
	$S$ & Discrete state space with values $\{1,...,J\}$ and observations are denoted as $s$\\
	$\bar{\bP}$ & $J \times J$ passive transition probability matrix\\
	$\bar{p}_{ij}$ & An element of $\bar{\bP}$; passive transition probability from state $i$ to state $j$\\
	$\gamma$ & Discount factor in $[0,1]$\\
	$R$ & State cost function with values denoted $r_i$ for $i \in S$\\
	$\bu$ & Continous controls which define the policy \eqref{eq:controlprob} \\
	$u_{ij}$ & An element of $\bu$ \\
	$p_{ij}(u_{ij})$ & Controlled transitions or policy defined by continuous controls and passive dynamics \eqref{eq:controlprob}\\
	$p^*(s_t = j \vert s_i = i)$ & Same as $p_{ij}(u_{ij})$\\
	$l(\cdot, \bu)$ & State and control cost function; it is the sum of the state cost $R$ and KL divergence between passive and controlled transition probabilities \eqref{eq:costs}\\
	$\bv$ & Cost-to-go function or the expected discounted future state control costs \eqref{eq:CTG} with values denoted by $v_i$ for $i \in S$\\	
\end{tabulary}

\end{appendix}

%%%%%%%%%%%%%%%%%%%%%%%%%%%%%%%%%%%%%%%%%%%%%%
%% Support information (funding), if any,   %%
%% should be provided in the                %%
%% Acknowledgements section.                %%
%%%%%%%%%%%%%%%%%%%%%%%%%%%%%%%%%%%%%%%%%%%%%%
\section*{Acknowledgements}
This material is based upon work supported by the National Science Foundation Graduate Research Fellowship Program under Grant No. 1443129. Any opinions, findings, and conclusions or recommendations expressed in this material do not necessarily reflect the views of the National Science Foundation. CKW was supported by NSF Grant DMS-1811745; MBH was supported by NSF Grant DMS-1614392. Any use of trade, firm, or product names is for descriptive purposes only and does not imply endorsement by the U.S. Government.
 
%%%%%%%%%%%%%%%%%%%%%%%%%%%%%%%%%%%%%%%%%%%%%%
%% Supplementary Material, if any, should   %%
%% be provided in {supplement} environment  %%
%% with title inside \textbf{} and short    %%
%% description below.                       %%
%%%%%%%%%%%%%%%%%%%%%%%%%%%%%%%%%%%%%%%%%%%%%%
\begin{supplement}
\stitle{Stan Algorithms and Code}. 
\slink[doi]{COMPLETED BY THE TYPESETTER}
\sdatatype{.pdf}
\sdescription{Definitions of the HMC, NUTS, and variational approximation algorithms and Stan model code.}
\end{supplement}
\begin{supplement}
\stitle{Guppy movement animations}. 
\slink[doi]{COMPLETED BY THE TYPESETTER}
\sdatatype{.zip}
\sdescription{Animations of the observed movement for one guppy experiment along with simulated trajectories from the learned policy: stochastic and least cost behavior.}
\end{supplement}

%%  imsart-nameyear.bst  will be used to                   %%
%%  MR numbers will be added by VTeX.                      %%
%%                                                         %%
%%  Use \cite{...} to cite references in text.             %%
%%                                                         %%
%%%%%%%%%%%%%%%%%%%%%%%%%%%%%%%%%%%%%%%%%%%%%%%%%%%%%%%%%%%%%

%% if your bibliography is in bibtex format, uncomment commands:
\nocite{schafer2020bayesian}
\bibliographystyle{imsart-nameyear} % Style BST file
\bibliography{IRL.bib}       % Bibliography file (usually '*.bib')

%% or include bibliography directly:
% \begin{thebibliography}{}
% \bibitem[\protect\citeauthoryear{???}{???}]{b1}
% \end{thebibliography}

\end{document}

%\section{}
%\subsection{}

\section*{Supplement A: STAN Algorithms and Code}\label{appendix3:stan}

We implemented the MCMC sampling and variational approximation for Bayesian inference using the modeling program STAN \citep{carpenter2017stan}. 

The default MCMC sampling algorithm used by STAN is the Hamiltonian Monte Carlo (HMC) with no-U-Turn sampler (NUTS). The HMC algorithm is an efficient algorithm for proposing parameters for exploring the posterior distribution \citep{betancourt2015hamiltonian}. Then, the proposed parameters are accepted in a Metropolis update.  The parameters are proposed by moving along the gradient of the conditional posterior and the size of the move is determined by auxiliary momentum variables. Inbetween proposals, the moves are repeated a fixed number of times (leapfrog steps) and the final value of the parameters are evaluated for acceptance. The NUTS provides automatic tuning of the number of leapfrog steps needed for adequate exploration of the posterior \citep{hoffman2014nuts}. We sampled four chains of 6000 iterations with the first 2000 discarded as burnin and checked diagnostics using the tools provided by the R package rstan \citep{stan2020}

Variational Bayesian inference is an optimization problem that minimizes the Kullback-Leibler divergence between an approximate posterior distribution and the true posterior distribution \citep{tzikas2008variational}. The true posterior distribution is unknown and thus, the KL divergence cannot be computed exactly but it can be bounded. This bound is referred to as the evidence lower bound (ELBO). The parameters of the approximate posterior distribution are updated iteratively using stochastic gradient ascent until the change in the ELBO is sufficiently small. STAN implements the algorithm of \citet{kucukelbir2015automatic} and stops when the change in ELBO is less than 0.01.

\newpage

\subsection{STAN Model Code}
The following code is used for both the MCMC sampling and variational inference with STAN.

   \lstset{language=C++,
           basicstyle=\ttfamily,
          breaklines=true
          }

\begin{lstlisting}
data {
  int<lower=1> N; // number of states
  int<lower=1> Nv; // number of v parameters
  int<lower=1> Ni; // number of individuals
  int<lower=1> Nobs; // number of observations of n_ijk
  int<lower=1> Nb; //number of basis functions
  int<lower=1> maxcount; // max frequency of counts
  real<lower=0, upper=1> discount; // MDP discount factor
  int<lower=0,upper=maxcount> n[Nobs]; // vector of counts 
  int<lower=1,upper=N> J[Nobs]; // vector of from state indices
  int<lower=1,upper=N> K[Nobs]; // vector of to state indices
  int<lower=1,upper=Ni> I[Nobs]; // vector of individual indices
  matrix[Nv,Nb] Fmat; // matrix of features, if identity than slopes model
  row_vector<lower=0,upper=1>[N] Aw[Nv]; //adjacency matrix of transition indices map for w
}
parameters {
  vector[Nb] w; // z parameters
  real<lower=0> alpha; // v precision
}
transformed parameters{
  vector[Nv] v;
  v = Fmat*w;
}
model {
  real tmp;
  alpha ~ gamma(0.1,0.1);
  w ~ normal(0,1/alpha);  
  for(i in 1:Nobs){
    tmp = 0;
    // loop over the neighboring states    
    tmp = Aw[J[i],]*exp(-discount*v);
    target += n[i]*((-discount*v[K[i]])-log(tmp));
  }
}
\end{lstlisting}

\newpage

\bibliographystyle{imsart-nameyear} % Style BST file
%\bibliography{IRL}       % Bibliography file (usually '*.bib')